\documentclass[letterpaper, 10 pt, conference]{ieeeconf}  

\IEEEoverridecommandlockouts                              

\usepackage{amssymb} 

\usepackage{amsthm}

\usepackage{graphicx}
\usepackage{makecell}
\usepackage{threeparttable} 
\usepackage{subcaption}
\usepackage{xcolor}
\usepackage{booktabs}
\usepackage{multirow}

\usepackage[pagebackref = false,breaklinks,colorlinks,citecolor=green]{hyperref}

\usepackage{xurl}

\newif\ifproofread
\proofreadfalse

\title{\LARGE \bf
Improving Monocular Visual-Inertial Initialization with Structureless Visual-Inertial Bundle Adjustment
}

\author{Junlin Song, Antoine Richard, and Miguel Olivares-Mendez
\thanks{Space Robotics (SpaceR) Research Group, Int. Centre for Security, Reliability and Trust (SnT), University of Luxembourg, Luxembourg.} 
}

\begin{document}

\maketitle
\thispagestyle{empty}
\pagestyle{empty}

\begin{abstract}

Monocular visual inertial odometry (VIO) has facilitated a wide range of real-time motion tracking applications, thanks to the small size of the sensor suite and low power consumption. To successfully bootstrap VIO algorithms, the initialization module is extremely important. Most initialization methods rely on the reconstruction of 3D visual point clouds. These methods suffer from high computational cost as state vector contains both motion states and 3D feature points. To address this issue, some researchers recently proposed a structureless initialization method, which can solve the initial state without recovering 3D structure. However, this method potentially compromises performance due to the decoupled estimation of rotation and translation, as well as linear constraints. To improve its accuracy, we propose novel structureless visual-inertial bundle adjustment to further refine previous structureless solution. Extensive experiments on real-world datasets show our method significantly improves the VIO initialization accuracy, while maintaining real-time performance.

\end{abstract}

\begin{keywords}
Visual-Inertial Initialization, Visual-Inertial Bundle Adjustment
\end{keywords}

\section{Introduction}

Visual inertial odometry (VIO) has gained great popularity thanks to its ability to  provide real-time, accurate 6 degree-of-freedom (DoF) motion tracking by fusing camera and IMU data. The success of VIO leans on the complementary characteristics of these two sensors. The camera perceives rich environmental information in the field of view, and is able to correct IMU biases. While the IMU measures high-frequency acceleration and angular velocity, enhancing robustness in fast motion (motion blur) and weak texture scenes. For the minimal sensor suite, i.e. monocular VIO, the IMU also overcomes the scale issue of monocular vision. Thanks to the small size of the sensor suite, low power consumption, and low cost, VIO is widely used in AR/VR \cite{Google, Apple, Meta, fan2024schurvins}, robotics \cite{wu2017vins, delmerico2018benchmark, kang2023view, song2024accurate}, and planetary exploration \cite{bayard2019vision, delaune2021range}.

Existing VIO algorithms can be broadly classified into two categories: optimization-based methods and filter-based methods. Optimization-based methods include OKVIS \cite{leutenegger2015keyframe}, VINS-Mono \cite{qin2018vins}, and ORB-SLAM3 \cite{campos2021orb}. Filter-based methods include ROVIO \cite{bloesch2017iterated}, and Multi-State Constraint Kalman Filter (MSCKF) \cite{mourikis2007multi, geneva2020openvins}. The first step for all these methods to successfully converge is to feed the VIO estimator with accurate initial states, typically including pose, velocity, and IMU biases. The pose and velocity are expressed in the gravity-aligned global frame.

Accuracy and efficiency are two key factors for the VIO initialization. On one hand, inaccurate VIO initialization has adverse effect on the convergence of the state, and can even lead to the divergence of the estimator. On the other hand, high latency will impact real-time performance. In early studies of VIO algorithms \cite{leutenegger2015keyframe, bloesch2017iterated, geneva2020openvins} the initial states are roughly recovered by assuming the motion is static. However, this assumption is not suitable for dynamic motion. VIO initialization caused by visual tracking loss or unhealthy estimator (abnormal covariance) during motion is very common in real-world use-cases. Therefore, precise and fast initialization in dynamic motion is highly desired. In this paper, we focus on dynamic monocular VIO initialization.

Numerous VIO initialization methods have been proposed to simultaneously recover the initial states and 3D environmental structure (visual landmarks) by solving visual-inertial structure from motion (VI-SfM) or visual-inertial bundle adjustment (VI-BA) \cite{qin2018vins, campos2021orb, geneva2020openvins, merrill2023fast}. The estimation of 3D structure typically requires large computational cost. To achieve higher computational efficiency and better accuracy, \cite{he2023rotation} first proposed a structureless initialization scheme that does not require the 3D structure reconstruction, and demonstrated superior performance over other state-of-the-art (SOTA) structure-based initialization methods. However, the decoupled estimation of rotation and translation, and the formulation of linear translation constraints can limit its performance. In order to fully exploit all available sensor measurements and further explore accuracy improvement, we propose a novel structureless VI-BA to refine the initial solution from \cite{he2023rotation}, without sacrificing structureless characteristics. Our key contributions are summarized as:

\begin{itemize}
\item To the best of our knowledge, this is the first structureless VI-BA. This novel VI-BA framework utilizes IMU preintegration constraints and visual epipolar constraints to optimize the initial VIO states without the reconstruction of 3D visual point clouds, significantly reducing the dimension of state variables.

\item We apply structureless VI-BA to tackle the monocular VIO initialization problem. Extensive experimental results show initialization accuracy can be greatly improved by structureless VI-BA, while maintaining superior computational efficiency.

\end{itemize}

\section{Related work}

Structure-based monocular VIO initialization methods can be divided into two categories, based on whether IMU measurements are used for the initialization of visual trajectory and structure (3D feature points). If IMU measurements are not needed for structure initialization, it is classified as loosely-coupled, otherwise it is tightly-coupled.

Typical loosely-coupled methods include \cite{qin2017robust, qin2018vins, campos2020inertial, campos2021orb}. Pure visual SfM is leveraged to obtain up-to-scale camera trajectory and structure expressed in the camera frame \cite{qin2017robust, qin2018vins}. By aligning the rotation motions of IMU and camera, the gyroscope bias is solved. Then, the translation motions of IMU and camera are correlated and formulated as a linear system to be solved for the remaining initial states. \cite{campos2020inertial, campos2021orb} get the initial camera trajectory through pure visual SLAM. Then, the motion correlation between IMU and camera is formulated as a maximum a posteriori estimation problem considering measurement uncertainty. The initial states are solved iteratively. Finally, structure-based VI-BA is performed to optimize both the initial state and structure. One drawback of loosely-coupled methods is that they rely on pure visual SfM or SLAM at the beginning, without considering the accuracy or robustness improvement from complementary IMU measurements on the initial trajectory, especially in challenging scenarios such as rapid rotation.

To jointly use IMU and visual bearing measurements for structure initialization, some researchers propose tightly-coupled methods, including \cite{martinelli2014closed, evangelidis2021revisiting, dong2012estimator, geneva2022openvins}. The gyroscope bias is assumed to be known or even negligible to obtain the rotational motion estimation. Furthermore, utilizing measurements from two sensors, the initial states and structure are jointly estimated through closed-form solutions. However, without accounting for gyroscope bias, the initialization accuracy could be severely affected \cite{kaiser2016simultaneous}. Meanwhile, the joint estimation for the visual landmarks may result in unstable numerical solution due to the lack of prior knowledge about the 3D information of feature points. \cite{zhou2022learned, merrill2023fast, merrill2024fast} attempt to use monocular depth prior inferred from deep learning based approach to better constrain initialization problem. However, these methods may suffer from the generalization issue inherited from deep learning and demand GPU enabled devices, significantly increasing computing resource and cost, and reducing power efficiency.

Both the loosely-coupled and tightly-coupled method mentioned above need to estimate the 3D visual structure. Processes like SfM, SLAM, or structure-based VI-BA are typically computationally expensive. Is it possible to tackle the VIO initialization problem without recovering the structure? Few papers address it from this perspective.

\cite{he2023rotation} take the first step to investigate the structure-less method. The rotation constraints from the IMU \cite{forster2016manifold} are directly integrated into the frame-to-frame rotation constraints \cite{kneip2013direct}, which makes the gyroscope bias estimation gain benefit from tightly-coupled mode and improves rotation motion estimations. Then, both loosely-coupled and tightly-couple solutions are proposed in \cite{he2023rotation} to estimate the rest of the initial states. These two solutions are formulated with linear translation constraints, and do not require the 3D structure reconstruction. Extensive experiments in \cite{he2023rotation} show this structure-less initialization method outperforms SOTA structure-based initialization methods \cite{qin2018vins, geneva2022openvins}, in terms of accuracy and computational efficiency. However, decoupled estimation and linear constraints can compromise accuracy performance. To fully explore the accuracy of structure-less initialization method, we propose a novel structure-less VI-BA to refine the initial solution obtained from \cite{he2023rotation}.

\section{Notation}

The IMU frame, the camera frame, and the gravity-aligned global frame are expressed as $\left\{ I \right\}$, $\left\{ C \right\}$, and $\left\{ G \right\}$, respectively. The 3D position of a point, $f$, is represented as ${}^A{p_f}$ in a rigid coordinate frame $\left\{ A \right\}$. We use quaternions to represent the rotation of rigid bodies, which is a non-singular expression. For detailed introduction and properties of quaternion, we refer to \cite{sola2017quaternion}. The orientation of frame $\left\{ B \right\}$ relative to frame $\left\{ A \right\}$ is represented by ${}_A^Bq$, and its corresponding rotation matrix is given by ${}_A^BR$. The coordinate frame transformation is represented as
\begin{equation}
    {}^B{p_f} = {}_A^BR{}^A{p_f} + {}^B{p_A}
\end{equation}

Quaternion algebra is different from linear algebra. The addition of quaternion and vector, as well as the subtraction between two quaternions, is noted as
\begin{equation} \label{eq:quat}
  \begin{array}{l}
    q \boxplus \delta : = q \otimes \exp \left( {\frac{\delta }{2}} \right)\\
    {q_1} \boxminus {q_2}: = 2\log \left( {q_2^{ - 1} \otimes {q_1}} \right)
  \end{array}
\end{equation}

Where $\exp \left( \bullet \right)$ maps a vector in tangent space to quaternion. Logarithmic mapping performs the opposite operation. $\delta $ is a vector in tangent space. $ \otimes $ represents quaternion multiplication. Equation (\ref{eq:quat}) is typically employed to formulate orientation incremental update and measurement residual.

${\left[  \bullet  \right]_ \times }$ and $\left\|  \bullet  \right\|$ denote the skew symmetric matrix and Euclidean norm corresponding to a 3D vector, respectively. ${\left[  \bullet  \right]^T}$ is used to represent the transpose of a matrix. 

\section{Proposed Initialization Framework}

\begin{figure}[htbp]
    \centering
    \includegraphics[width=0.4\textwidth]{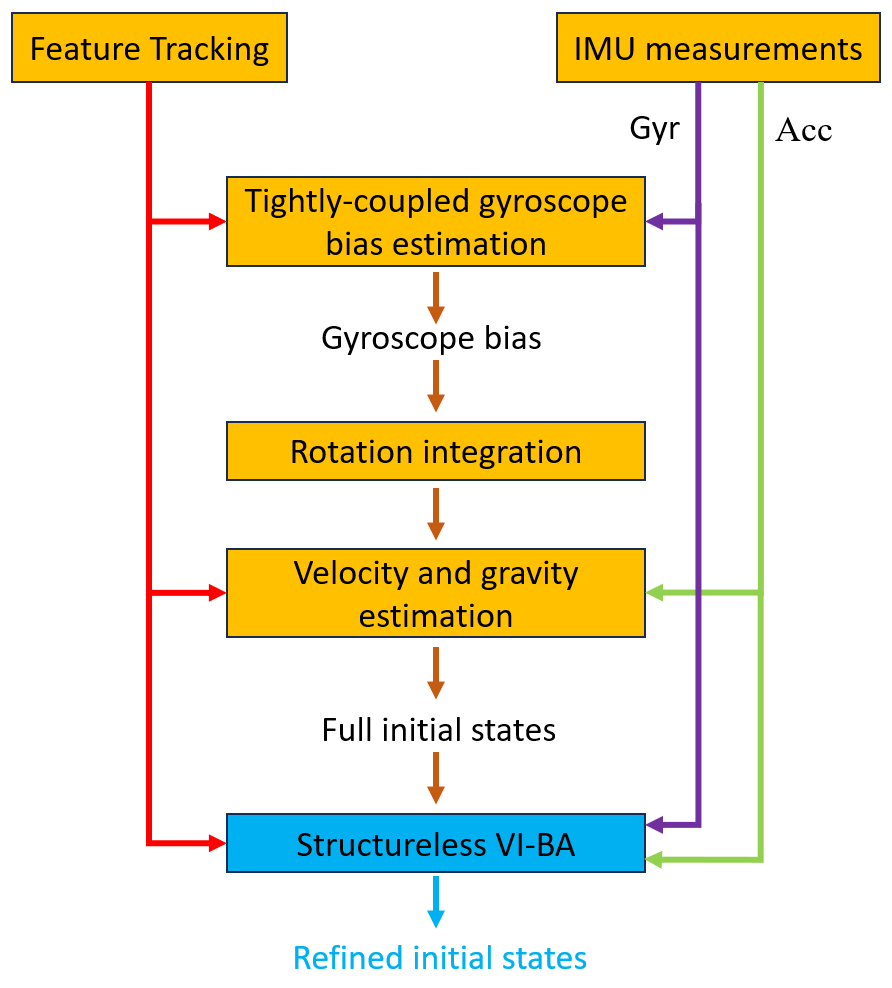}
    \caption{Structureless VIO Initialization framework enhanced by structureless VI-BA (highlighted in blue).}
    \label{framework}
\end{figure}

The structureless monocular VIO initialization method described in \cite{he2023rotation} outperforms SOTA structure-based initialization methods \cite{qin2018vins, geneva2022openvins}, in terms of accuracy and computational efficiency. Its accuracy advantage stems from tightly-coupled gyroscope bias estimation method, while its efficiency advantage benefits from structureless characteristics, as traditional triangulation and depth estimation for feature points are no longer compulsory. Taking inspiration from \cite{he2023rotation}, we propose ``structureless VI-BA'' to further refine the initial solution. The overall framework is shown in the Fig. \ref{framework}, and our novel contribution is highlighted in blue.

\section{Visual-Inertial Bundle Adjustment}

\begin{figure}[htbp]
  \centering
    \begin{subfigure}[t]{0.23\textwidth}
        \centering
        \includegraphics[width=\textwidth, height=\textwidth]{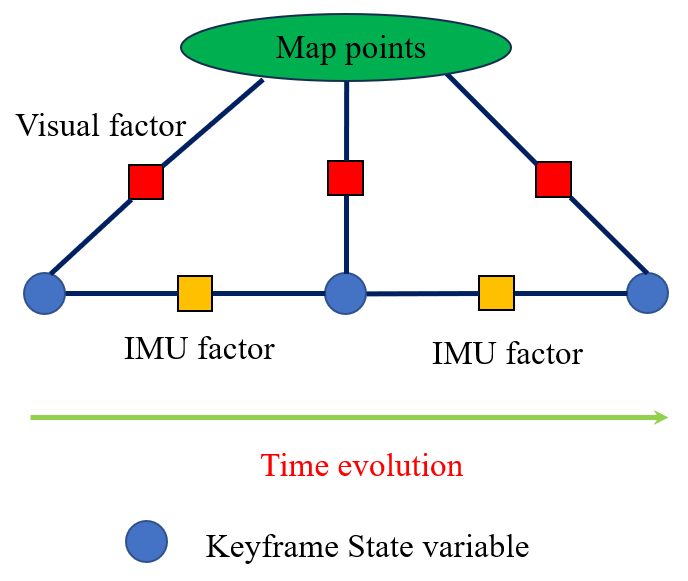}
        \label{factor1}
    \end{subfigure}
    \hfill
    \begin{subfigure}[t]{0.23\textwidth}
        \centering
        \includegraphics[width=\textwidth, height=\textwidth]{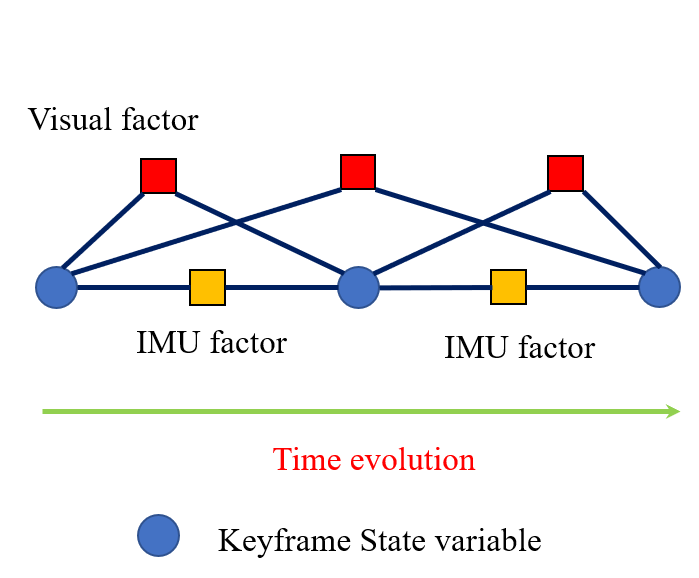}
        \label{factor2}
    \end{subfigure}
  \caption{Left: factor graph for structure-based VI-BA. Right: factor graph for structureless VI-BA.}
  \label{factor}
\end{figure}

The main purpose of this section is to review the classic structure-based VI-BA and introduce the novel structureless VI-BA. Structure-based VI-BA requires an initial visual map. Instead, structureless VI-BA does not perform the 3D structure reconstruction, and the visual constraints are obtained via epipolar geometry. Factor graphs are presented in the Fig. \ref{factor}.

\subsection{Structure-based VI-BA}

Structure-based VI-BA typically refines the following state
\begin{equation} \label{eq:sb}
    \begin{array}{l}
    x = {\left[ {\begin{array}{*{20}{c}}
    {x_{{c_0}}^T}& \cdots &{x_{{c_N}}^T}&{{}^Gp_{{f_1}}^T}& \cdots &{{}^Gp_{{f_M}}^T}
    \end{array}} \right]^T}\\
    x_{{c_k}}^T = {\left[ {\begin{array}{*{20}{c}}
    {{}^Gp_{{I_k}}^T}&{{}^Gv_{{I_k}}^T}&{{}_{{I_k}}^G{q^T}}&{b_{{a_k}}^T}&{b_{{g_k}}^T}
    \end{array}} \right]^T}
    \end{array}
\end{equation}

Where state vector $x$ includes $N + 1$ keyframe states and $M$ environmental features observed by these keyframes. ${x_{{c_k}}}$ represents the IMU state at keyframe timestamp ${t_k}$, including the IMU's position ${}^G{p_{{I_k}}}$, velocity ${}^G{v_{{I_k}}}$, orientation ${}_{{I_k}}^Gq$, accelerometer bias ${b_{{a_k}}}$, and gyroscope bias ${b_{{g_k}}}$. ${}^G{p_{{f_l}}}$ denotes the 3D position of environmental feature point $f_l$. Although some researchers prefer others feature point parameterization, such as inverse depth parameterization, we only use basic Euclidean parameterization to illustrate the structure-based VI-BA here. For more information on feature parameterization, we refer interested readers to OpenVINS's documentation \cite{update-feat}.

The goal of structure-based VI-BA is to optimize all keyframe states and environmental feature points, given their prior estimation values. Available constraints include IMU measurements between keyframes and observations of feature points by keyframes. The structure-based VI-BA can be formulated as the following optimization problem
\begin{equation}
    \mathop {\min }\limits_x \left\{ {\sum\limits_{k = 1}^N {\left\| {{r_{{I_{k - 1,k}}}}} \right\|_{{\Sigma _{{I_{k - 1,k}}}}}^2}  + \sum\limits_{l = 1}^M {\sum\limits_{i \in {{\rm K}^l}} {{\rho _{Hub}}\left( {\left\| {{r_{il}}} \right\|_{{\Sigma _C}}^2} \right)} } } \right\}
\end{equation}

Where ${r_{{I_{k - 1,k}}}}\left( {{x_{{c_{k - 1}}}},{x_{{c_k}}}} \right)$ and ${r_{il}}$ are the IMU preintegration residual and visual reprojection residual, respectively. Their detailed definitions are provided in Section \ref{Sec:IMU} and Section \ref{Sec:Reprojection}. ${K^l}$ is the set of keyframes observing feature point ${f_l}$. A robust Huber kernel function ${\rho _{Hub}}\left(  \bullet  \right)$ is used to mitigate the impact of pixel observation outliers.

\subsubsection{IMU preintegration residual} \label{Sec:IMU}

High-frequency IMU measurements between adjacent keyframes are typically formulized as preintegration terms \cite{forster2016manifold} to reduce computational cost caused by repeated integration during multiple non-linear optimization iterations. Given the accelerometer bias and gyroscope bias, preintegrated measurements of position, velocity, and rotation are
\begin{equation}
    \begin{array}{l}
    \alpha _k^{k - 1} = \int {\int_{{t_{k - 1}}}^{{t_k}} {{}_{{I_\tau }}^{{I_{k - 1}}}R\left( {{a_\tau } - {b_{{a_{k - 1}}}}} \right)} } d{\tau ^2}\\
    \beta _k^{k - 1} = \int_{{t_{k - 1}}}^{{t_k}} {{}_{{I_\tau }}^{{I_{k - 1}}}R\left( {{a_\tau } - {b_{{a_{k - 1}}}}} \right)} d\tau \\
    \gamma _k^{k - 1} = \int_{{t_{k - 1}}}^{{t_k}} {\frac{1}{2}{}_{{I_\tau }}^{{I_{k - 1}}}q \otimes \left( {{\omega _\tau } - {b_{{g_{k - 1}}}}} \right)} d\tau 
    \end{array}
\end{equation}

Where ${t_{k - 1}}$ and ${t_k}$ represent the timestamp of the ($k-1$)-th keyframe and the $k$-th keyframe, respectively. ${a_\tau }$ and ${\omega _\tau }$ represent the raw accelerometer and gyroscope measurements between ${t_{k - 1}}$ and ${t_k}$. It is noticed that preintegration measurements depend on the given IMU biases $\left\{ {{b_{{a_{k - 1}}}},{b_{{g_{k - 1}}}}} \right\}$. If the estimation of bias changes, preintegrated measurements should also be adjusted accordingly. This dependency can be addressed through a first-order Taylor expansion
\begin{equation}
    \begin{array}{l}
    \alpha _k^{k - 1} \approx \alpha _k^{k - 1} + J_{{b_a}}^\alpha \delta {b_{{a_{k - 1}}}} + J_{{b_g}}^\alpha \delta {b_{{g_{k - 1}}}}\\
    \beta _k^{k - 1} \approx \beta _k^{k - 1} + J_{{b_a}}^\beta \delta {b_{{a_{k - 1}}}} + J_{{b_g}}^\beta \delta {b_{{g_{k - 1}}}}\\
    \gamma _k^{k - 1} \approx \gamma _k^{k - 1} \otimes \left[ {\begin{array}{*{20}{c}}
    1\\
    {\frac{1}{2}J_{{b_g}}^\gamma \delta {b_{{g_{k - 1}}}}}
    \end{array}} \right]
    \end{array}
\end{equation}

The final residuals formulated by the preintegrated IMU measurements between two adjacent keyframe states can be defined as
\begin{equation}
    \begin{array}{l}
    {r_{{I_{k - 1,k}}}}\left( {{x_{{c_{k - 1}}}},{x_{{c_k}}}} \right) = \left[ {\begin{array}{*{20}{c}}
    {\alpha _k^{k - 1} - \bar \alpha _k^{k - 1}}\\
    {\beta _k^{k - 1} - \bar \beta _k^{k - 1}}\\
    {\gamma _k^{k - 1} \boxminus \bar \gamma _k^{k - 1}}\\
    {{b_{{a_k}}} - {b_{{a_{k - 1}}}}}\\
    {{b_{{g_k}}} - {b_{{g_{k - 1}}}}}
    \end{array}} \right]\\
    \bar \alpha _k^{k - 1} = {}_G^{{I_{k - 1}}}R\left( {{}^G{p_{{I_k}}} - {}^G{p_{{I_{k - 1}}}} - {}^G{v_{{I_{k - 1}}}}dt - \frac{1}{2}gd{t^2}} \right)\\
    \bar \beta _k^{k - 1} = {}_G^{{I_{k - 1}}}R\left( {{}^G{v_{{I_k}}} - {}^G{v_{{I_{k - 1}}}} - gdt} \right)\\
    \bar \gamma _k^{k - 1} = {}_G^{{I_{k - 1}}}q \otimes {}_{{I_k}}^Gq
    \end{array}
\end{equation}

Where $g$ is the gravity vector and $dt$ denotes the time interval from ${t_{k - 1}}$ to ${t_k}$.

The IMU preintegration measurements in this subsection are illustrated with a quaternion-based derivation. For the Jacobians of residuals with respect to state variables and measurement covariance ${\Sigma _{{I_{k - 1,k}}}}$, we refer readers to \cite{qin2018vins}.  For other forms of IMU preintegration, interested readers are referred to \cite{forster2016manifold, eckenhoff2019closed}.

\subsubsection{Visual reprojection residual} \label{Sec:Reprojection}

The preintegration residuals in the previous subsection describe the efficient IMU constraint on adjacent two keyframes. As visual features are tracked across multiple consecutive keyframes, visual constraints can be built on multiple keyframes. If a feature point ${f_l}$ is observed by the $i$-th keyframe, the measurement model of a feature point can be written as the classic reprojection residual, given the 3D position of ${f_l}$ and the pose of the $i$-th keyframe
\begin{equation}
    \begin{array}{l}
    {r_{il}} = {u_{il}} - {h_d}\left( {z_{il}^n,\zeta } \right)\\
    z_{il}^n = {h_p}\left( {{}^{{C_i}}{p_{{f_l}}}} \right)\\
    {}^{{C_i}}{p_{{f_l}}} = {h_t}\left( {{}^G{p_{{f_l}}},{}_G^{{C_i}}R,{}^G{p_{{C_i}}}} \right)
    \end{array}
\end{equation}

Where ${u_{il}}$ is the raw uv pixel observation. ${h_d}\left( { \bullet , \bullet } \right)$ represents the camera distortion equation, and $\zeta $ is the camera intrinsic parameter. ${h_p}\left(  \bullet  \right)$ represents the camera projection equation, converting the 3D position of a feature point in the camera frame into normalized coordinates, also known as feature bearing. ${h_t}\left(  \bullet  \right)$ represents the coordinate transformation equation, converting the 3D position of a feature point in the global frame into the camera frame. The camera pose corresponding to the $i$-th keyframe is $\left\{ {{}_G^{{C_i}}R,{}^G{p_{{C_i}}}} \right\}$, associating IMU pose with the following equation
\begin{equation}
    \begin{array}{l}
    {}_{{C_i}}^GR = {}_{{I_i}}^GR{}_C^IR\\
    {}^G{p_{{C_i}}} = {}^G{p_{{I_i}}} + {}_{{I_i}}^GR{}^I{p_C}
    \end{array}
\end{equation}

Where $\left\{ {{}_C^IR,{}^I{p_C}} \right\}$ are the extrinsic parameters between IMU and camera, which can be calibrated together with the camera intrinsic parameter $\zeta $ before BA. Note that these calibration parameters can also be refined in the VI-BA. The corresponding measurement covariance of ${r_{il}}$ is typically set by engineering experience, such as fixed at 1 pixel. For more information on the Jacobians, we also refer interested readers to OpenVINS documentation \cite{update-feat}.

Up until now, we present the complete classic structure-based VI-BA. One characteristic of this BA is the need to recover 3D structure of the scene, which requires large computational effort, especially when numerous feature points are involved in state vector (\ref{eq:sb}).

\subsection{Structureless VI-BA}

Structureless VIO initialization \cite{he2023rotation} shows better accuracy and efficiency compared to structure-based VIO initialization \cite{qin2018vins, geneva2022openvins}. However, decoupled estimation of rotation and translation can limit initialization accuracy. Therefore, we propose novel structureless VI-BA to optimize the initial states obtained from \cite{he2023rotation}, without sacrificing structureless characteristics.

\begin{figure}[htbp]
    \centering
    \includegraphics[width=0.3\textwidth]{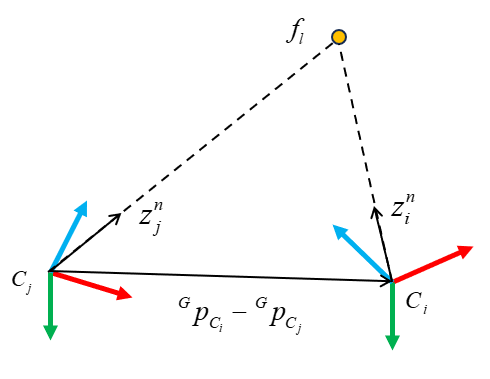}
    \caption{Co-planar geometric relationships for feature bearing vectors with the frame-to-frame translation vector.}
    \label{epipolar}
\end{figure}

State variables of structureless VI-BA can be obtained by deleting the environmental feature points in Equation (\ref{eq:sb})
\begin{equation}
    \begin{array}{l}
    x = {\left[ {\begin{array}{*{20}{c}}
    {x_{{c_0}}^T}& \cdots &{x_{{c_N}}^T}
    \end{array}} \right]^T}\\
    x_{{c_k}}^T = {\left[ {\begin{array}{*{20}{c}}
    {{}^Gp_{{I_k}}^T}&{{}^Gv_{{I_k}}^T}&{{}_{{I_k}}^G{q^T}}&{b_{{a_k}}^T}&{b_{{g_k}}^T}
    \end{array}} \right]^T}
    \end{array}
\end{equation}

The purpose of structureless VI-BA is to refine the IMU states for all keyframes, given their initial values. Unlike structure-based VI-BA, visual measurements are formulated by epipolar constraints, eliminating the dependence on 3D structure completely. Structureless VI-BA can be expressed as the following optimization problem
\begin{equation} \label{eq:opt}
    \scalebox{1.0}{$
    \mathop {\min }\limits_x \left\{ {\sum\limits_{k = 1}^N {\left\| {{r_{{I_{k - 1,k}}}}} \right\|_{{\Sigma _{{I_{k - 1,k}}}}}^2}  + \sum\limits_{l = 1}^M {\sum\limits_{\left( {i,j} \right) \in {{\rm K}^l}} {{\rho _{Hub}}\left( {\left\| {r_{ij}^n} \right\|_{{\Sigma _C}}^2} \right)} } } \right\}
    $}
\end{equation}

Where ${r_{{I_{k - 1,k}}}}\left( {{x_{{c_{k - 1}}}},{x_{{c_k}}}} \right)$ is the IMU preintegration residual as defined in Section \ref{Sec:IMU}. While $r_{ij}^n$ is the visual measurement residual generated by epipolar geometry. If a keyframe pair $\left( {i,j} \right)$ observes the same feature point ${f_l}$, this co-view relationships can be expressed using epipolar constraints, given the poses and normalized coordinates observations of the keyframe pair
\begin{equation}
    \begin{array}{l}
    z_i^n = h_d^{ - 1}\left( {{u_i},\zeta } \right),
    {\quad\quad} z_j^n = h_d^{ - 1}\left( {{u_j},\zeta } \right)
    \end{array}
\end{equation}

Where $u_i$ and $u_j$ represent the raw uv pixel observations of the same feature point ${f_l}$ from the keyframe pair $\left( {i,j} \right)$. The back projection function of the camera is denoted as $h_d^{ - 1}\left( { \bullet , \bullet } \right)$, which converts the raw uv pixel observation into normalized coordinates using the camera intrinsic parameter $\zeta $. The visual measurement residual generated by the epipolar constraint can be represented as
\begin{equation}
    \begin{array}{l}
    r_{ij}^n\left( {{x_{{c_i}}},{x_{{c_j}}}} \right) = {\left( {{}_{{I_j}}^GR{}_C^IRz_j^n} \right)^T}{\left[ {\frac{t}{{\left\| t \right\|}}} \right]_ \times }\left( {{}_{{I_i}}^GR{}_C^IRz_i^n} \right)\\
    t = {}^G{p_{{C_i}}} - {}^G{p_{{C_j}}}\\
     = {}^G{p_{{I_i}}} + {}_{{I_i}}^GR{}^I{p_C} - {}^G{p_{{I_j}}} - {}_{{I_j}}^GR{}^I{p_C}
    \end{array}
\end{equation}

Where $r_{ij}^n$ has intuitive geometric interpretation, i.e., two feature bearing vectors should be co-planar with the frame-to-frame translation vector $t$, as depicted in the Fig. \ref{epipolar}. All direction vectors are expressed in the global frame. $t$ is normalized to prevent it from converging to 0. 

Next, we discuss the analytic Jacobians of the residual with respect to related state variables. Defining the following intermediate variables
\begin{equation}
    \begin{array}{l}
    r_{ij}^n\left( {{x_{{c_i}}},{x_{{c_j}}}} \right) = {A^T}{\left[ C \right]_ \times }B\\
    A = {}_{{I_j}}^GR{}_C^IRz_j^n,
    {\quad\quad} C = \frac{t}{{\left\| t \right\|}},
    {\quad\quad} B = {}_{{I_i}}^GR{}_C^IRz_i^n
    \end{array}
\end{equation}

The Jacobians of residual with respect to $A$, $C$, and $B$ are
\begin{equation}
    \begin{array}{l}
    \frac{{\partial r_{ij}^n}}{{\partial A}} = {\left( {{{\left[ C \right]}_ \times }B} \right)^T}\\
    \frac{{\partial r_{ij}^n}}{{\partial C}} =  - {A^T}{\left[ B \right]_ \times }\\
    \frac{{\partial r_{ij}^n}}{{\partial B}} = {A^T}{\left[ C \right]_ \times }
    \end{array}
\end{equation}

The Jacobians of $A$, and $B$ with respect to rotation are
\begin{equation}
    \begin{array}{l}
    \frac{{\partial A}}{{\partial {}_{{I_j}}^GR}} =  - {}_{{I_j}}^GR{\left[ {{}_C^IRz_j^n} \right]_ \times }\\
    \frac{{\partial B}}{{\partial {}_{{I_i}}^GR}} =  - {}_{{I_i}}^GR{\left[ {{}_C^IRz_i^n} \right]_ \times }
    \end{array}
\end{equation}

The Jacobians of $t$ with respect to pose are calculated as
\begin{equation}
    \begin{array}{l}
    t = {}^G{p_{{I_i}}} + {}_{{I_i}}^GR{}^I{p_C} - {}^G{p_{{I_j}}} - {}_{{I_j}}^GR{}^I{p_C}\\
    \frac{{\partial t}}{{\partial {}^G{p_{{I_i}}}}} = {I_3},
    {\quad\quad} \frac{{\partial t}}{{\partial {}_{{I_i}}^GR}} =  - {}_{{I_i}}^GR{\left[ {{}^I{p_C}} \right]_ \times }\\
    \frac{{\partial t}}{{\partial {}^G{p_{{I_j}}}}} =  - {I_3},
    {\quad\quad} \frac{{\partial t}}{{\partial {}_{{I_j}}^GR}} = {}_{{I_j}}^GR{\left[ {{}^I{p_C}} \right]_ \times }
    \end{array}
\end{equation}

The Jacobian of $C$ with respect to $t$ is
\begin{equation}
    \frac{{\partial C}}{{\partial t}} = \left[ {\begin{array}{*{20}{c}}
    {\frac{1}{{\left\| t \right\|}} - \frac{{t_x^2}}{{{{\left\| t \right\|}^3}}}}&{\frac{{ - {t_x}{t_y}}}{{{{\left\| t \right\|}^3}}}}&{\frac{{ - {t_x}{t_z}}}{{{{\left\| t \right\|}^3}}}}\\
    {\frac{{ - {t_x}{t_y}}}{{{{\left\| t \right\|}^3}}}}&{\frac{1}{{\left\| t \right\|}} - \frac{{t_y^2}}{{{{\left\| t \right\|}^3}}}}&{\frac{{ - {t_y}{t_z}}}{{{{\left\| t \right\|}^3}}}}\\
    {\frac{{ - {t_x}{t_z}}}{{{{\left\| t \right\|}^3}}}}&{\frac{{ - {t_y}{t_z}}}{{{{\left\| t \right\|}^3}}}}&{\frac{1}{{\left\| t \right\|}} - \frac{{t_z^2}}{{{{\left\| t \right\|}^3}}}}
    \end{array}} \right]
\end{equation}

According to the chain rule, the Jacobians of residual $r_{ij}^n$ with respect to the poses of keyframe pair $\left( {i,j} \right)$ can be computed as
\begin{equation}
    \begin{array}{l}
    \frac{{\partial r_{ij}^n}}{{\partial {}_{{I_i}}^GR}} = \frac{{\partial r_{ij}^n}}{{\partial B}}\frac{{\partial B}}{{\partial {}_{{I_i}}^GR}} + \frac{{\partial r_{ij}^n}}{{\partial C}}\frac{{\partial C}}{{\partial t}}\frac{{\partial t}}{{\partial {}_{{I_i}}^GR}}\\
    \frac{{\partial r_{ij}^n}}{{\partial {}^G{p_{{I_i}}}}} = \frac{{\partial r_{ij}^n}}{{\partial C}}\frac{{\partial C}}{{\partial t}}\frac{{\partial t}}{{\partial {}^G{p_{{I_i}}}}}\\
    \frac{{\partial r_{ij}^n}}{{\partial {}_{{I_j}}^GR}} = \frac{{\partial r_{ij}^n}}{{\partial A}}\frac{{\partial A}}{{\partial {}_{{I_j}}^GR}} + \frac{{\partial r_{ij}^n}}{{\partial C}}\frac{{\partial C}}{{\partial t}}\frac{{\partial t}}{{\partial {}_{{I_j}}^GR}}\\
    \frac{{\partial r_{ij}^n}}{{\partial {}^G{p_{{I_j}}}}} = \frac{{\partial r_{ij}^n}}{{\partial C}}\frac{{\partial C}}{{\partial t}}\frac{{\partial t}}{{\partial {}^G{p_{{I_j}}}}}
    \end{array}
\end{equation}

Finally, the non-linear least square optimization problem derived by structureless VI-BA (\ref{eq:opt}) can be tackled with specific solver, like ceres-solver\footnote{\url{http://ceres-solver.org}}.

\section{Results}

\begin{table*}
\caption{Exhaustive initialization results on the EuRoC Dataset. The initialization sliding window size is set to 10.}
\label{table_euroc}
\begin{center}
\begin{tabular}{|c|ccc|ccc|ccc|ccc|}
\hline
\multirow{2}{*}{\makecell{Sequence}} & \multicolumn{3}{c|}{ATE (m)} & \multicolumn{3}{c|}{ATE (deg)}  & \multicolumn{3}{c|}{Velocity RMSE (m/s)} & \multicolumn{3}{c|}{Avg solve time (ms)}\\
\cline{2-13}   & \cite{qin2018vins} & \cite{he2023rotation} & Ours & \cite{qin2018vins} & \cite{he2023rotation} & Ours & \cite{qin2018vins} & \cite{he2023rotation} & Ours & \cite{qin2018vins} & \cite{he2023rotation} & Ours\\
\hline
MH\_01\_easy & 0.046 & 0.046  & \textbf{0.027} & 0.101 & 0.084  & \textbf{0.031} & 0.167 & 0.132 & \textbf{0.076} & 67.95 & \textbf{8.61} & 36.73\\
MH\_02\_easy & 0.039 & 0.041  & \textbf{0.019} & 0.128 & 0.068  & \textbf{0.031} & 0.124 & 0.126 & \textbf{0.063} & 65.61 & \textbf{8.58} & 36.12\\
MH\_03\_medium & 0.143 & 0.185  & \textbf{0.132} & 0.182 & 0.125  & \textbf{0.033} & 0.451 & 0.358 & \textbf{0.199} & 57.96 & \textbf{8.46} & 34.69\\
MH\_04\_difficult & 0.179 & 0.131  & \textbf{0.054} & 0.173 & 0.114  & \textbf{0.051} & 0.549 & 0.388 & \textbf{0.164} & 66.76 & \textbf{8.34} & 35.57\\
MH\_05\_difficult & 0.162 & 0.134  & \textbf{0.057} & 0.178 & 0.138  & \textbf{0.030} & 0.495 & 0.382 & \textbf{0.154} & 63.65 & \textbf{8.50} & 34.77\\
\hline\hline
V1\_01\_easy & 0.060 & 0.045  & \textbf{0.021} & 0.213 & 0.131  & \textbf{0.080} & 0.185 & 0.143 & \textbf{0.075} & 55.68 & \textbf{8.43} & 35.76\\
V1\_02\_medium & 0.160 & 0.098  & \textbf{0.042} & 0.554 & 0.225  & \textbf{0.134} & 0.500 & 0.314 & \textbf{0.147} & 54.33 & \textbf{7.83} & 28.25\\
V1\_03\_difficult & 0.146 & 0.106  & \textbf{0.069} & 0.605 & 0.380  & \textbf{0.245} & 0.459 & 0.327 & \textbf{0.238} & 54.32 & \textbf{7.24} & 24.15\\
V2\_01\_easy & 0.042 & 0.031  & \textbf{0.015} & 0.228 & 0.171  & \textbf{0.089} & 0.132 & 0.099 & \textbf{0.054} & 62.42 & \textbf{8.75} & 35.12\\
V2\_02\_medium & 0.084 & 0.054  & \textbf{0.027} & 0.390 & 0.158  & \textbf{0.082} & 0.260 & 0.171 & \textbf{0.097} & 54.58 & \textbf{7.67} & 28.07\\
V2\_03\_difficult & 0.177 & 0.145  & \textbf{0.087} & 1.001 & 0.491  & \textbf{0.393} & 0.495 & 0.327 & \textbf{0.213} & 55.23 & \textbf{8.06} & 21.89\\
\hline\hline
Avg & 0.113 & 0.092 & \textbf{0.050} &  0.341 & 0.190 &  \textbf{0.109} & 0.347 & 0.252 & \textbf{0.135} & 59.86 &
  \textbf{8.22} & 31.92\\
\hline
\end{tabular}
\end{center}
\end{table*}

\begin{table*}
\caption{Exhaustive initialization results on the TUM-VI Dataset. The initialization sliding window size is set to 10.}
\label{table_tum}
\begin{center}
\begin{tabular}{|c|ccc|ccc|ccc|ccc|}
\hline
\multirow{2}{*}{\makecell{Sequence}} & \multicolumn{3}{c|}{ATE (m)} & \multicolumn{3}{c|}{ATE (deg)}  & \multicolumn{3}{c|}{Velocity RMSE (m/s)} & \multicolumn{3}{c|}{Avg solve time (ms)}\\
\cline{2-13}   & \cite{qin2018vins} & \cite{he2023rotation} & Ours & \cite{qin2018vins} & \cite{he2023rotation} & Ours & \cite{qin2018vins} & \cite{he2023rotation} & Ours & \cite{qin2018vins} & \cite{he2023rotation} & Ours\\
\hline
room1 & 0.106 & 0.162  & \textbf{0.063} & 0.985 & 0.579  & \textbf{0.521} & 0.162 & 0.230 & \textbf{0.125} & 84.60 & \textbf{12.18} & 30.16\\
room2 & 0.270 & 0.184  & \textbf{0.065} & 1.195 & 0.712  & \textbf{0.621} & 0.436 & 0.298 & \textbf{0.140} & 85.68 & \textbf{12.47} & 32.44\\
room3 & 0.143 & 0.239  & \textbf{0.091} & 1.035 & 0.715  & \textbf{0.647} & 0.224 & 0.373 & \textbf{0.173} & 105.33 & \textbf{12.14} & 29.93\\
room4 & 0.097 & 0.117  & \textbf{0.060} & 1.069 & 0.581  & \textbf{0.556} & 0.141 & 0.159 & \textbf{0.128} & 79.44 & \textbf{13.16} & 33.54\\
room5 & 0.120 & 0.197  & \textbf{0.101} & 1.241 & 0.708  & \textbf{0.668} & 0.225 & 0.251 & \textbf{0.207} & 105.95 & \textbf{11.99} & 28.26\\
room6 & 0.147 & 0.071  & \textbf{0.021} & 0.985 & 0.415  & \textbf{0.347} & 0.220 & 0.111 & \textbf{0.044} & 76.98 & \textbf{12.34} & 34.90\\
\hline\hline
corridor1 & 0.078 & 0.100  & \textbf{0.048} & 0.950 & 0.604  & \textbf{0.542} & 0.135 & 0.141 & \textbf{0.080} & 96.80 & \textbf{11.69} & 29.66\\
corridor2 & 0.132 & 0.107  & \textbf{0.035} & 1.202 & 0.687  & \textbf{0.619} & 0.206 & 0.140 & \textbf{0.057} & 81.11 & \textbf{13.03} & 33.99\\
corridor3 & 0.122 & 0.137  & \textbf{0.059} & 1.390 & 0.658  & \textbf{0.584} & 0.180 & 0.191 & \textbf{0.117} & 86.63 & \textbf{11.87} & 31.82\\
corridor4 & 0.120 & 0.076  & \textbf{0.029} & 1.113 & 0.609  & \textbf{0.485} & 0.206 & 0.110 & \textbf{0.048} & 83.93 & \textbf{11.61} & 29.92\\
corridor5 & 0.108 & 0.107  & \textbf{0.061} & 1.603 & 0.560  & \textbf{0.530} & 0.159 & 0.168 & \textbf{0.120} & 98.47 & \textbf{12.75} & 36.60\\
\hline\hline
Avg & 0.131 & 0.136 & \textbf{0.058} & 1.161 & 0.612 & \textbf{0.556} & 0.209 & 0.197 & \textbf{0.113} & 89.54 &
 \textbf{12.29} & 31.93\\
\hline
\end{tabular}
\end{center}
\end{table*}

Our proposed initialization method is compared with two baseline methods, including VINS-Mono \cite{qin2018vins} and DRT-l. DRT-l is a strong baseline recently proposed by \cite{he2023rotation}, and experiments in \cite{he2023rotation} have shown its superior performance compared to VINS-Mono, ov\_init \cite{geneva2022openvins}, and DRT-t. DRT-t is another structureless initialization solution proposed by \cite{he2023rotation}. VINS-Mono is one of the most popular optimization-based VIO algorithms. The initialization module of VINS-Mono is a loosely-coupled structure-based initialization algorithm. VINS-Mono exhibits better initialization performance than ov\_init, and weaker performance than DRT-l. We introduce VINS-Mono as additional baseline to demonstrate the real-time performance of our proposed method.

For fair comparison, all algorithms adopt the same feature tracking front-end and keyframe selection scheme from VINS-Mono. For the setting of front-end related parameters, IMU noise, and visual residual covariance, we also refer to VINS-Mono\footnote{\url{https://github.com/HKUST-Aerial-Robotics/VINS-Mono/tree/master/config}}. To verify the performance of different initialization algorithms, we employ two popular VIO datasets, EuRoC \cite{burri2016euroc} and TUM-VI \cite{schubert2018tum}. All the experiments are conducted on a laptop computer with an Intel(R) Xeon(R) W-10855M CPU @ 2.80GHz, and 16 GB of RAM.

We maintain an initialization sliding window with a fixed number of keyframes and exhaustively perform initialization for each sequence. Each time the latest keyframe of the initialization window moves forward, i.e., when the latest received image is selected as a new keyframe, we repeatedly perform initialization. This exhaustive initialization test takes into account that VIO may be requested to be initialized at any time due to unexpected hardware or software anomalies, or unhealthy trend of the estimator\footnote{\url{https://science.nasa.gov/blog/surviving-an-in-flight-anomaly-what-happened-on-ingenuitys-sixth-flight}}.


Absolute trajectory error (ATE) \cite{zhang2018tutorial} and velocity norm error are utilized as accuracy metrics, while initialization solve time is recorded as efficiency metric. ATE can be obtained by aligning the initialization trajectory with the groundtruth trajectory in posyaw mode. The velocity norm error is obtained by calculating the RMSE for the initialized velocity norm and the groundtruth velocity norm over the entire initialization window. Assuming the initialization solve time for one specific window is ${t_i}$, we calculate the average solve time over the entire sequence as the final metric 
\begin{equation}
    \bar t = \frac{1}{N_S}\sum\limits_{i = 1}^{N_S} {{t_i}} 
\end{equation}

Where $N_S$ indicates how many initializations we perform on a certain sequence. The accuracy metrics are also averaged over the entire sequence.

\subsection{EuRoC dataset}

The EuRoC dataset is collected by a micro air vehicle (MAV), and includes hardware-synchronized stereo image and IMU data, which are available at 20Hz and 200Hz respectively. Our experiments only use the left camera image. The EuRoC dataset contains 11 sequences, which are classified as easy, medium, and difficult based on several factors, like motion velocity and illumination variation.

We evaluated accuracy and efficiency for VINS-Mono, DRT-l, and our proposed method on all 11 sequences. Results are summarized in TABLE \ref{table_euroc}. DRT-l outperforms VINS-Mono in both accuracy and efficiency, which is consistent with the conclusion in \cite{he2023rotation}. On average, our method further improves accuracy compared to DRT-l, with 45.7\% reduction in translation ATE, 42.6\% reduction in rotation ATE, and 46.4\% reduction in velocity RMSE. This is expected, as the structureless VI-BA employed in our method can fully exploit all available visual and IMU measurements. The decoupled estimation and linear constraints in \cite{he2023rotation} lead to potential information loss, thereby compromising accuracy.

Although the average solve time is increased due to the introduction of structureless VI-BA, our method still achieves real-time performance. Our average solve time is 31.92ms, which is less than the time gap (50ms) between two consecutive frames. While the average solve time of VINS-Mono is 59.86ms, greater than 50ms. High computational cost of VINS-Mono mainly attributes to the expensive SfM process to recover 3D structure. Results demonstrate the structureless initialization scheme brings remarkable computational efficiency advantage over structure-based initialization. This advantage allows us to use structureless VI-BA to perform optimization while maintaining real-time performance.

\subsection{TUM-VI dataset}

\begin{figure}[htbp]
    \centering
    \includegraphics[width=0.48\textwidth]{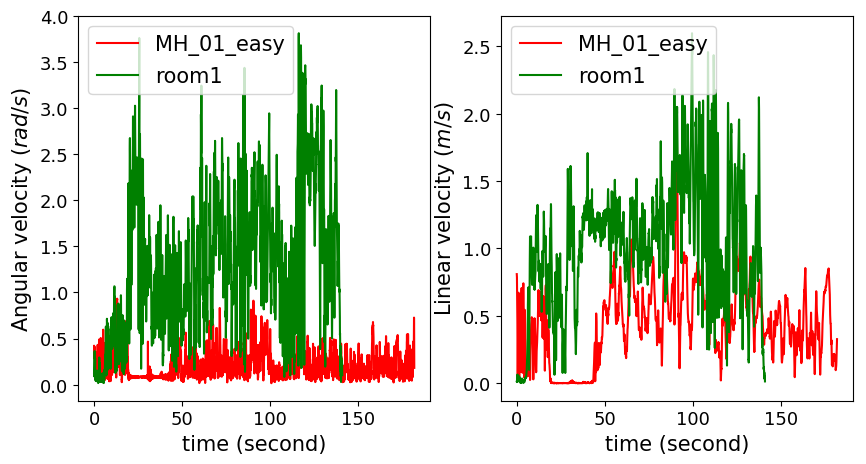}
    \caption{Angular and linear velocity profiles of MH\_01\_easy from EuRoC dataset and room1 from TUM-VI dataset.}
    \label{velocity}
\end{figure}

The TUM-VI dataset is collected with a handheld device, receiving image and IMU data at 20Hz and 200Hz, respectively. Our experiments also only use images from the left fisheye camera. The accuracy and efficiency metrics are reported in TABLE \ref{table_tum}. Results again confirm that our method can further improve accuracy over DRT-l while maintaining real-time performance compared to VINS-Mono. We can observe that the rotation accuracy of all three methods has notable decrease compared to the EuRoC dataset, which can be attributed to the much more aggressive motion of the TUM-VI dataset, as shown in the Fig. \ref{velocity}.

The translation ATE and velocity RMSE for TUM-VI dataset are comparable to those of EuRoC dataset, possibly because more feature tracks in the capture scenes from TUM-VI dataset can provide more sufficient constraints, which explains the longer solve time of VINS-Mono and DRT-l.

\section{Conclusion}

In this paper, we introduce novel structureless VI-BA to improve the  accuracy of recently proposed structureless monocular VIO initialization method \cite{he2023rotation}. To maintain the structureless characteristics, this novel VI-BA adopts epipolar constraint instead of reprojection error to formulate visual measurement. Experimental results on two publicly benchmark datasets demonstrate that our method outperforms SOTA methods in accuracy, while still maintaining competitive computational efficiency. The efficient nature of structureless scheme allows us to utilize limited computing resource to perform more refinement operations. 
Proposed complete structureless VIO initialization framework can be further improved. For future work, it is interesting to explore if we can find better structureless visual constraint on multiple views to upgrade epipolar constraint on two views.









\bibliographystyle{ieeetr}
\bibliography{bib}

\end{document}